%% file: main.tex
\documentclass[10pt,twocolumn,letterpaper]{article}

\usepackage[pagenumbers]{cvpr} 

\input{preamble}

\usepackage{hwemoji}
\usepackage{cuted}   
\usepackage{capt-of} 
\definecolor{cvprblue}{rgb}{0.21,0.49,0.74}
\usepackage{hyperref}
\hypersetup{colorlinks,allcolors=cvprblue}
\hypersetup{pagebackref,breaklinks}

\usepackage{xcolor,colortbl}

\definecolor{bestcolor}{HTML}{FFCCCC}
\definecolor{secondcolor}{HTML}{FFE5B4}
\definecolor{thirdcolor}{HTML}{FFFACD}

\def\paperID{4622} 
\def\confName{CVPR}
\def\confYear{2026}

\title{⚡ FlashMesh: Faster and Better Autoregressive Mesh Synthesis\\ via Structured Speculation}

\author{Tingrui Shen\thanks{Equal Contribution}\\
South China University of Technology
\and
Yiheng Zhang\footnotemark[1]\\
Tsinghua University
\and
Chen Tang\footnotemark[1]\\
South China University of Technology
\and
Chuan Ping\\
Zhejiang University
\and
Zixing Zhao\\
Tencent VISVISE
\and
Le Wan\\
Tencent VISVISE
\and
Yuwang Wang\\
Tsinghua University
\and
Ronggang Wang\\
Peking University
\and
Shengfeng He\thanks{Corresponding author: shengfenghe@smu.edu.sg}\\
Singapore Management University
}

\begin{document}
\input{pics/sec1_pic_teaser}
\maketitle
\input{sec/0_abstract}
\input{sec/1_intro}
\input{sec/2_related_work}

\input{sec/4_method}

\input{sec/5_exp}
\input{sec/6_conclusion}

{
    \small
    \bibliographystyle{ieeenat_fullname}
    \bibliography{main}
}
\input{sec/X_suppl}


\end{document}

%% file: preamble.tex
\usepackage{float}



%% file: pics/sec1_pic_teaser.tex
\teaser{
    \centering
    \includegraphics[width=1.0\linewidth, trim=50 170 0 70, clip]{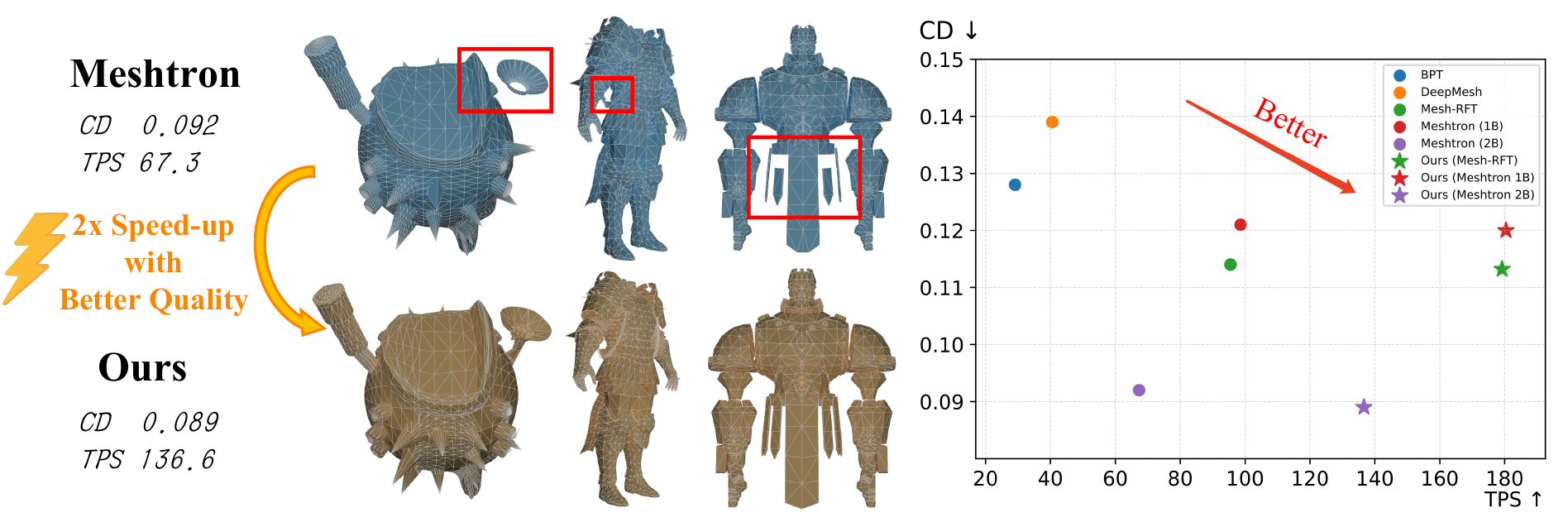}
    \caption{We propose \textit{FlashMesh}, an efficient and high-quality autoregressive mesh generation framework. Compared to the baseline Meshtron, FlashMesh produces visually superior results with \textit{2× faster inference}. Quantitative comparisons on the right, using Chamfer Distance (CD) and Tokens per Second (TPS), show consistent improvements over multiple baselines.}
    \label{fig:teaser}
} 

%% file: sec/0_abstract.tex
\begin{abstract}
Autoregressive models can generate high-quality 3D meshes by sequentially producing vertices and faces, but their token-by-token decoding results in slow inference, limiting practical use in interactive and large-scale applications.
We present FlashMesh, a fast and high-fidelity mesh generation framework that rethinks autoregressive decoding through a \textit{predict-correct-verify} paradigm. The key insight is that mesh tokens exhibit strong structural and geometric correlations that enable confident multi-token speculation. FlashMesh leverages this by introducing a speculative decoding scheme tailored to the commonly used hourglass transformer architecture, enabling parallel prediction across face, point, and coordinate levels.
Extensive experiments show that FlashMesh achieves up to a 2$\times$ speedup over standard autoregressive models while also improving generation fidelity. Our results demonstrate that structural priors in mesh data can be systematically harnessed to accelerate and enhance autoregressive generation.
\end{abstract} 

%% file: sec/1_intro.tex
\section{Introduction}
\label{sec:intro}

High-quality 3D mesh generation is a fundamental task in virtual reality~\cite{martinez2020unrealrox, bahirat2018designing}, gaming~\cite{chornyi2025development}, and digital content creation~\cite{chen2024deep, jiang2024survey}. Meshes offer a compact, editable representation of geometry, supporting fine surface detail, efficient rendering, and direct topological control. Unlike dense formats such as voxels~\cite{zhou20213d, schwarz2022voxgraf} or implicit fields~\cite{chen2019learning, erkocc2023hyperdiffusion}, meshes represent surfaces explicitly through vertices and faces, making them well-suited for high-fidelity and resource-efficient applications. As a result, learning-based mesh generation has become an increasingly important focus within 3D vision and graphics.

Recent autoregressive models~\cite{siddiqui2024meshgpt, zhang2025vertexregen, lionar2025treemeshgpt, wang2025iflame, chen2025xspecmesh} have shown promising results by generating mesh elements (e.g., vertices, faces, and coordinates) token by token, capturing complex topology and geometric structure with high accuracy. However, their strictly sequential decoding process creates a fundamental bottleneck: each token must be predicted before the next can be generated, leading to long inference times. This inefficiency severely limits their usability in interactive scenarios or large-scale pipelines where speed is critical.

To overcome this bottleneck, we revisit the decoding process itself. Speculative decoding strategies~\cite{gloeckle2024better, liu2024deepseek} in large language models provide a promising direction: a lightweight draft model predicts multiple tokens in parallel, and a main model verifies them. Adapting this idea to mesh generation, however, is far from straightforward. Mesh data are not flat sequences but hierarchical structures composed of faces, vertices, and coordinates, each with strong geometric and topological dependencies. Moreover, modern mesh generation models typically use Hourglass Transformer architectures that compress and expand these hierarchical features, which differ fundamentally from the flat transformer decoders used in text models. These architectural and structural differences imply that speculative decoding must be redesigned to align with the unique properties of mesh representations. 

Guided by this observation, we propose \textit{FlashMesh}, a fast and high-quality autoregressive mesh generation framework based on a \textit{predict-correct-verify} paradigm. The core rationale of FlashMesh is that hierarchical mesh data contains predictable structural patterns that allow the model to confidently speculate multiple future tokens, provided that the speculative process respects geometric consistency and the architecture's feature hierarchy. FlashMesh operationalizes this idea through three coordinated components.

In the \textit{predict} stage, we introduce a speculative decoding strategy tailored to the hourglass architecture. Two lightweight modules, the SP-Block and HF-Block, exploit hierarchical feature compression to generate multiple future tokens in parallel. The SP-Block predicts future tokens from current-layer features, and the HF-Block incorporates information from coarser levels to refine these predictions and maintain hierarchical coherence.

In the \textit{correct} stage, we incorporate mesh-specific structural priors to resolve inconsistencies that arise from speculative prediction. A correction algorithm enforces vertex-sharing consistency among adjacent faces and adjusts geometrically incoherent coordinate predictions.

In the \textit{verify} stage, the backbone network evaluates corrected tokens in a single forward pass, ensuring that the final outputs remain faithful to the underlying autoregressive model.

Through this design, FlashMesh enables parallel generation of multiple tokens per step while preserving the fidelity advantages of autoregressive modeling. The framework leverages structural constraints rather than ignoring them, which leads to both faster inference and improved geometric and topological quality (see Fig.~\ref{fig:teaser}). FlashMesh represents a principled step toward scalable and accessible 3D mesh generation. 

In summary, our contributions are fourfold:

\begin{itemize}

\item We introduce FlashMesh, a novel autoregressive mesh generation framework based on a \textit{predict-correct-verify} paradigm for efficient and high-quality synthesis.

\item We propose a hierarchical speculative decoding strategy tailored to Hourglass Transformers, enabling parallel token prediction across face, point, and coordinate levels.

\item We develop a structure-aware correction module that enforces vertex-sharing consistency by leveraging mesh connectivity priors.

\item Extensive experiments show that FlashMesh achieves up to a 2$\times$ speedup over baselines while improving generation fidelity, advancing efficient autoregressive mesh modeling.

\end{itemize}

%% file: sec/2_related_work.tex
\section{Related work}
\label{sec:related_work}



\paragraph{Autoregressive Mesh Generation.}
Autoregressive models have recently shown strong performance in mesh generation by modeling the sequential distribution of faces, vertices, and coordinates directly from handcrafted datasets.  Meshtron~\citep{hao2024meshtron} introduced an Hourglass Transformer that factorizes the generation process into hierarchical face, vertex, and coordinate streams.
Building on this foundation, DeepMesh~\citep{zhao2025deepmesh} and Mesh-RFT~\citep{liu2025mesh} adopt reinforcement learning to enhance generation quality. ARMesh~\citep{lei2025armesh} and VertexRegen~\citep{zhang2025vertexregen} generate meshes in a partial-to-complete manner. Other methods such as BPT~\citep{weng2025scaling}, TreeMeshGPT~\citep{lionar2025treemeshgpt}, MeshAnything v2~\citep{chen2024meshanything}, EdgeRunner~\citep{tang2024edgerunner}, and MeshSilksong~\citep{song2025mesh} propose various token compression techniques to reduce sequence length.
Despite these advances, the sequential nature of autoregressive decoding remains a bottleneck. Since each token depends on the previous context, inference becomes slow, limiting deployment in large-scale applications. Addressing this limitation requires rethinking the decoding process itself.

\input{pics/sec3_whole_model.tex}

\paragraph{Efficient Decoding in Autoregressive Models.}
To accelerate autoregressive inference, recent work in language modeling has explored speculative and Jacobi-style decoding. Speculative decoding~\citep{chen2023accelerating, liu2023online, spector2023accelerating} uses a lightweight draft model to propose multiple tokens in parallel, followed by verification through a stronger main model. Jacobi-based approaches~\citep{santilli2023accelerating, teng2024accelerating, fu2024break, kou2024cllms} generate multiple tokens simultaneously and iteratively refine them for consistency.
In mesh generation, Iflame~\citep{wang2025iflame} proposes an interleaved decoding scheme, while XSpecMesh~\citep{chen2025xspecmesh} applies speculative decoding with LoRA-tuned draft models. These methods primarily focus on speed and often sacrifice geometric fidelity or structural consistency.
Addressing these limitations, FlashMesh builds on speculative decoding by introducing a predict–correct–verify paradigm designed specifically for hierarchical mesh representations. It incorporates structure-aware correction and token verification aligned with the hourglass architecture, enabling parallel decoding with better mesh quality.


%% file: pics/sec3_whole_model.tex
\begin{figure*}[t]
    \centering
    \includegraphics[width=\textwidth]{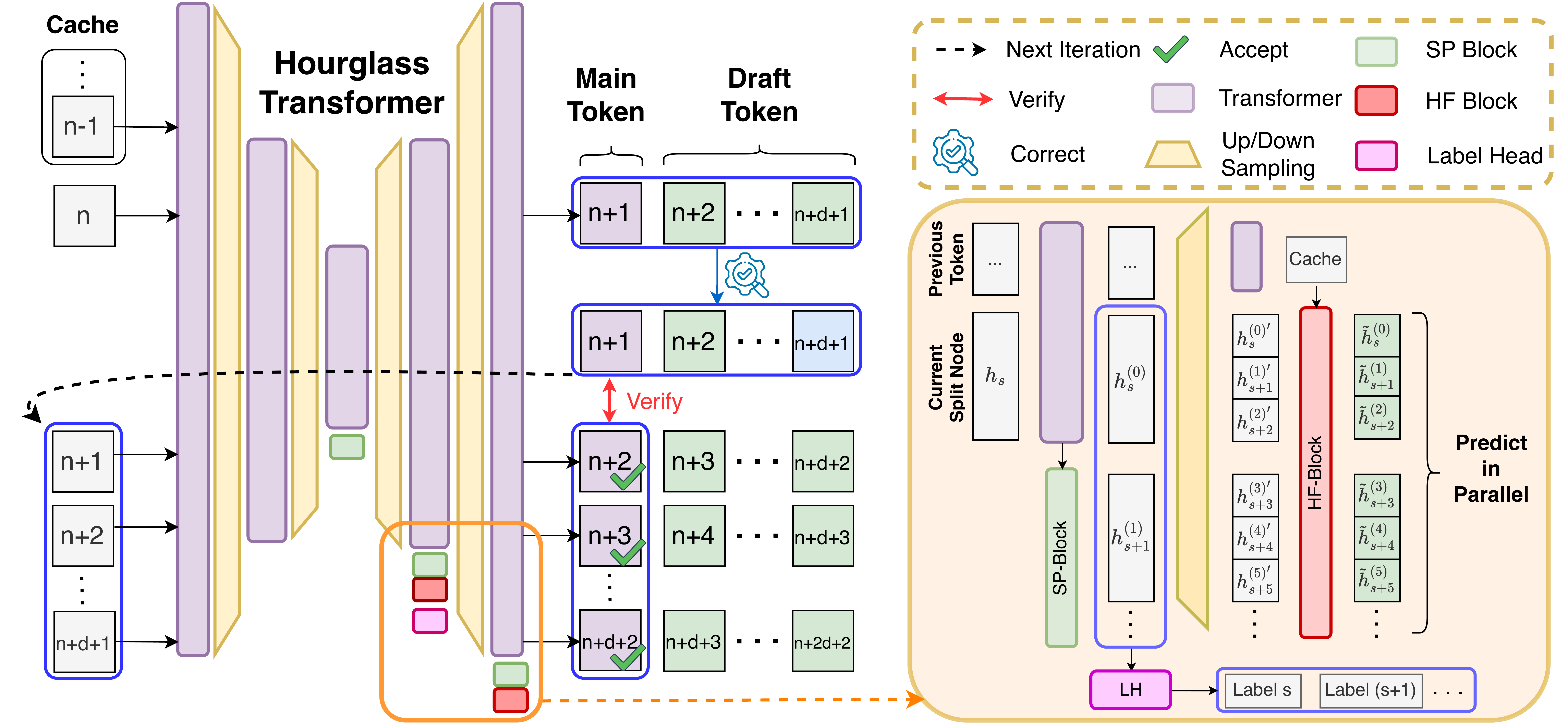}
    \vspace{-6mm}\caption{Overall architecture of the proposed predict-correct-verify framework. Predict: the original Hourglass Transformer generates main tokens, while the lightweight SP-Block and HF-Block parallelly produce draft tokens. Correct: a correction mechanism enforces vertex-sharing consistency. Verify: the backbone re-evaluates main and corrected draft tokens in a single forward pass and accepts the verified ones. Bottom right: Point-level pipeline of multi-layer multi-head speculative decoding.}
    \label{fig:whole_model}
\end{figure*} 

%% file: sec/4_method.tex
\section{Predict–Correct–Verify Paradigm}
\label{sec:pcv}

Autoregressive models are capable of generating high-quality meshes, yet their strictly sequential token-by-token inference severely limits generation speed. 
To overcome this bottleneck, we propose an efficient and high-quality \textbf{predict–correct–verify} framework for accelerating autoregressive mesh generation. 
This framework unifies parallel prediction and autoregressive precision by predicting multiple future tokens, correcting local geometric inconsistencies, and verifying the results progressively through the backbone network.
Our overall architecture is illustrated in~\cref{fig:whole_model}.

Specifically, the \textbf{Predict} stage (see Sec.~\ref{sec:speculative_decoding}) employs a Multi-Layer Multi-Head Speculative Decoding structure to predict multiple future tokens in parallel. In this stage, the backbone network (the original Hourglass Transformer) predicts the next token, referred to as the main token, while our proposed SP-Block and HF-Block collaboratively predict multiple subsequent draft tokens in parallel. Subsequently, the \textbf{Correct} stage (see Sec.~\ref{sec:correction}) performs point-level label prediction and geometric correction to ensure the local topological consistency of the simultaneously generated faces and vertices.
Finally, the \textbf{Verify} stage (see Sec.~\ref{sec:verify}) adopts a speculative decoding strategy, enabling the backbone to verify multiple draft tokens within a single forward pass, automatically discarding inconsistent predictions and significantly accelerating inference while enhancing generation quality.

Through the collaboration of these three stages, the predict–correct–verify framework achieves substantial generation acceleration while improving both efficiency and fidelity.
This paradigm not only reduces redundant autoregressive computation but also provides a new pathway toward efficient parallel generation of high-dimensional geometric structures.

\subsection{Multi-Layer Multi-Head Speculative Decoding}
\label{sec:speculative_decoding}

Building upon this paradigm, we introduce the multi-layer multi-head speculative decoding framework, which serves as the core of the predict stage. It is designed to accelerate autoregressive mesh generation by predicting multiple future tokens in parallel. Our design is based on an hourglass-style hierarchical decoder, which organizes mesh tokens across three levels—faces, vertices, and coordinates. The decoder first compresses coordinate-level embeddings into higher-level vertex and face embeddings to capture global geometry, and then progressively upsamples them back to low-level finer resolutions to recover local details. During this upsampling process, each transition node between levels (called \emph{split nodes}) expands coarse features into finer-grained mesh representations.



When the backbone reaches a split node, the SP-Block performs speculative prediction for the corresponding hierarchical features in parallel. The HF-Block then receives speculative hidden states from the upper level and interacts with the cached key–value states of the current layer. Through this hierarchical fusion, it integrates high-level structural cues with local context to produce the final features of the current layer. As illustrated in the bottom right of \cref{fig:whole_model}, we visualize a representative speculative decoding process that begins from the point-level split nodes. The original point features are processed by the SP-Block and then upsampled to obtain the predicted coordinate-level features, which are subsequently refined by the HF-Block to produce the final features. Similarly, speculative decoding from face-level split nodes passes through one SP-Block followed by two HF-Blocks to generate the final features, while speculative decoding from coordinate-level split nodes directly applies a single SP-Block to obtain the final features.

Each input coordinate token is processed by the backbone network to generate a \textit{main token}, and by three levels of SP-Blocks and HF-Blocks to produce multiple \textit{draft tokens}. Specifically, given an input token \(x_n\) at position \(n\), the backbone network outputs the main token \(x_{n+1}\), representing the predicted value of the next token. Meanwhile, the three levels of SP-Blocks and HF-Blocks simultaneously generate several draft token \(x_{n+2:n+D+1}\) representing the predicted value for positions \(n\!+\!2\) through \(n\!+\!D\!+\!1\), where $D$ denotes the number of the predicted draft tokens. Due to space limitations, we provide the detailed information about multi-layer multi-head speculative decoding (i.e. definition of the split node and the speculative decoding processes for the three levels) in the supplementary material.

\input{pics/sec4_pic_block}

\textbf{Speculative Prediction Block.} The speculative prediction block (SP-Block) is composed of multiple Transformer layers. Specifically, as illustrated in~\cref{fig:SPHF-Block} (a), the main backbone network preceding the SP-Block consists of $N-1$ transformer blocks, each containing a self-attention layer, a cross-attention layer for injecting the generation condition $c$, and a feed-forward network. Let $s$ denote the current token position, and define the input hidden state of the first layer as $h_s^{(0)} = \mathrm{Emb}(x_s)$, where $\mathrm{Emb}(\cdot)$ denotes the token embedding function and $x_s$ means the token at the position $s$. For each layer $l = 1, 2, \dots, N-1$, the output hidden state is computed as $h_s^{(l)} = \mathrm{Block}^{(l)}(h_s^{(l-1)}, c)$, and we denote the hidden state after the $(N-1)$-th layer as $h_s = h_s^{(N-1)}$. The backbone model then continues through the $N$-th layer to produce the final hidden state $h_s^{(N)}$. The SP-Block takes $h_s$ as input and processes it through $D$ parameter-independent transformer blocks, where the $d$-th decoding head predicts the token features at position $s+d$ as

\begin{equation}
h_{s+d}^{(d)} = \mathrm{Linear}\big( \mathrm{CA}^{(d)}(\mathrm{SA}^{(d)}(h_s), c) \big) + h_s,
\label{eq:spblock}
\end{equation}
where $\mathrm{SA}^{(d)}$ and $\mathrm{CA}^{(d)}$ denote the self-attention and cross-attention operations in the $d$-th speculative prediction block, respectively.

\textbf{Hierarchical Fusion Block.}
After speculative prediction by the SP-Block, the outputs \(h_{s+d}^{(d)}\) remain high-level representations and must interact with local, low-level context to yield accurate token predictions. To this end, we first convert each speculative high-level features into a short sequence of lower-level features via an upsampling module:
\begin{equation}
\big[ h_{s+3d}^{(3d)'},\; h_{s+3d+1}^{(3d+1)'},\; h_{s+3d+2}^{(3d+2)'} \big]
= \mathrm{Upsample}\!\big( h_{s+d}^{(d)} \big),
\label{eq:upsample}
\end{equation}
where \(\mathrm{Upsample}(\cdot)\) denotes the hourglass-style operator that expands a single high-level features into multiple finer-grained token features.

We then apply the Hierarchical Fusion Block (HF-Block) to let these upsampled features interact with current-layer key–value cache. Specifically, as shown in~\cref{fig:SPHF-Block} (b), for each upsampled features \(h_{s+t}^{(t)'}\) (where \(t\in\{3d, 3d+1, 3d+2\}\)), we compute a query and retrieve the shared cached keys and values:
\begin{align}
Q_{s+t}^{(t)} &= W_q^{(t)}\,h_{s+t}^{(t)'}, \\
K_{<s} &= W_k\,X^{k}_{<s}, \quad 
V_{<s} = W_v\,X^{v}_{<s},
\end{align}
where \(W_q^{(t)}\) is a layer-specific query projection for position \(t\), while \(W_k\) and \(W_v\) are shared linear projections applied to the kv-cache sequence \(X^{k}_{<s}\) and \(X^{v}_{<s}\) of prior token features produced by main backbone network.


Finally, the HF-Block produces the refined low-level features by applying an attention, an output projection, and a residual connection to the upsampled input:
\begin{equation}
\tilde{h}_{s+t}^{(t)} =  
h_{s+t}^{(t)'} + \mathrm{FFN}^{(t)}\!\Big(
\mathrm{Attn}\big(Q_{s+t}^{(t)}, K_{<s}, V_{<s}\big) 
\Big),
\label{eq:hf_output}
\end{equation}
where \(\mathrm{FFN}^{(t)}(\cdot)\) denotes a feed-forward network for position t.

\textbf{Loss Function.}
The coordinate prediction loss is defined as the average cross-entropy between the predicted token (i.e. main tokens and draft tokens) distributions and ground-truth tokensdistributions:
\begin{equation}
\mathcal{L}_{\mathrm{coord}} = - \frac{1}{N_c} \sum_{t=1}^{N_c} \log p_t(x_t),
\end{equation}
where $N_c$ means the total number of the predicted tokens and \(p_t(x_t)\) denotes the predicted probability of the ground-truth token \(x_t\) at position \(t\).

\subsection{Correction Mechanism}
\label{sec:correction}

\input{./pics/sec4_pic_correction}

When the model generates multiple faces in parallel, a fundamental issue arises:
when one face is being generated, the exact coordinates of the other faces within the same batch are still unknown.
As a result, adjacent faces that should share common vertices may instead produce misaligned points.

As shown in~\cref{fig:correction}, the blue triangles denote previously generated faces, while the red ones represent the newly generated batch.
Here, vertex~8 should coincide with vertex~6, and vertex~9 with vertex~3, but they are displaced.

To mitigate this issue, we attach a label head (i.e. a linear layer) after each point-level feature (see the bottom right of~\cref{fig:whole_model}) to classify every generated point into three categories:
(1) \emph{historical points}—coinciding with vertices generated in the previous batch;
(2) \emph{new points}—novel spatial positions; and
(3) \emph{intra-batch points}—duplicating new points generated earlier within the same batch.

For each intra-batch point, we first check for overlaps with other vertices in the same batch.
If none are found, we duplicate the nearest new point outside the current triangle to ensure local geometric consistency.
Finally, vertices are re-sorted along the $z$–$y$–$x$ axes to maintain the required autoregressive ordering. All these correction operations are applied only to the draft tokens, without modifying the main tokens. As illustrated in~\cref{fig:correction}, points~8 and~9 are corrected by duplicating points~6 and~3, respectively, followed by reordering according to their $z$-$y$-$x$ coordinates.

The label prediction head is supervised by a standard cross-entropy loss that encourages accurate classification of each point into the three categories described above. 
Formally, the label loss is defined as
\begin{equation}
\mathcal{L}_{\mathrm{label}} = - \frac{1}{N_p} \sum_{t=1}^{N_p} \log p_t(y_t),
\end{equation}
where $N_p$ means the total number of predicted label and \(p_t(y_t)\) denotes the predicted probability of the ground-truth label \(y_t\) for the \(t\)-th point.

The overall training objective combines the coordinate prediction loss and the label classification loss as
\begin{equation}
\mathcal{L}_{\mathrm{total}} = \mathcal{L}_{\mathrm{coord}} + \gamma \mathcal{L}_{\mathrm{label}},
\end{equation}
where \(\gamma\) is a balancing weight that controls the contribution of the label supervision.

\input{./pics/sec4_pic_verify.tex}
\subsection{Verify Mechanism}
\label{sec:verify}
After generating the next \(D\!+\!1\) tokens from position \(s\) using the backbone network, SP-Block and HF-Block, followed by the correction step, a straightforward approach would be to append these tokens to the existing sequence and continue prediction from position \(s\!+\!D\!+\!1\). 
However, due to the potential inaccuracy of the draft tokens, this strategy may degrade the overall sequence quality. 
To address this issue, we follow speculative decoding strategy~\cite{gloeckle2024better} from large language field, which enables the backbone model to verify multiple predicted tokens within a single forward pass.

Specifically, let \(s\) denote the position of the most recently accepted token. 
In the previous forward pass, the backbone predicted main token \(x_{s+1}\), while the SP-Block and HF-Block simultaneously predicted draft tokens \(x_{s+2:s+D+1}\). 
After applying the correction step, these tokens are fed into the next forward pass, where the backbone performs causal masking over \(x_{s+1:s+D+1}\) to obtain \(x'_{s+2:s+D+2}\). 
Since main token \(x_{s+1}\) was generated by the backbone itself, we directly accept this token. 
We then compare \(x_{s+2:s+D+1}\) with \(x'_{s+2:s+D+1}\) to identify the latest match, denoted as \(x_{s^{*}}\), and accept all tokens up to and including \(x_{s^{*}}\). 
Subsequently, the main token \(x_{s^{*}+1}\) together with the corresponding \(D\) draft tokens \(x_{s^{*}+2:s^{*}+D+1}\) are corrected and fed into the next forward pass. We illustrate our verification mechanism in~\cref{fig:verify}, where $D=2$ is used for clarity of presentation.

%% file: pics/sec4_pic_block.tex
\begin{figure}[t]
    \centering
    \includegraphics[width=\linewidth, trim=460 0 0 0, clip]{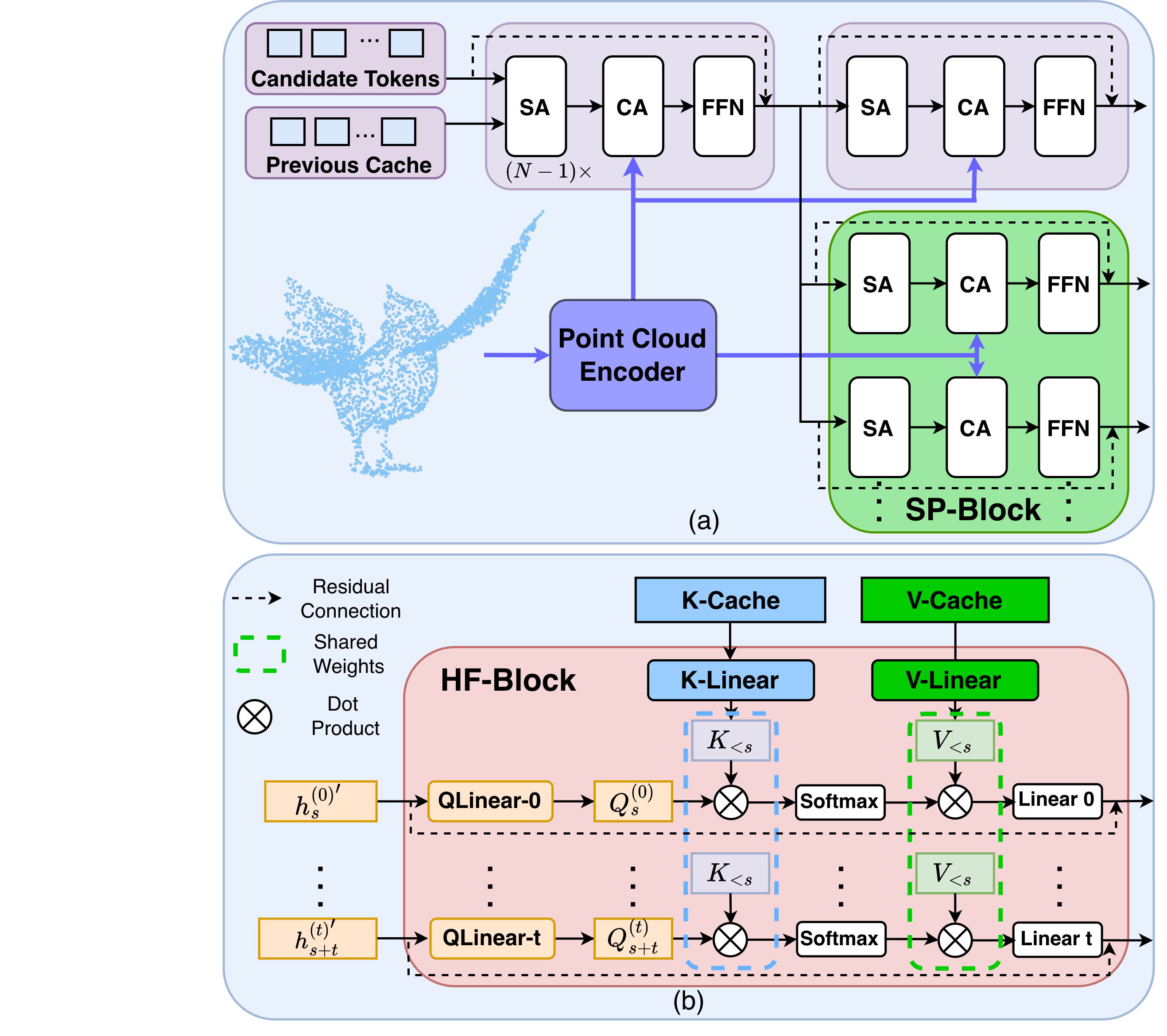}
    \vspace{-7mm}\caption{(a) The Speculative Prediction Block (SP-Block) predicts multiple draft tokens in parallel from the current hidden state.
(b) The Hierarchical Fusion Block (HF-Block) refines speculative embeddings by fusing them with cached local context for accurate token prediction.}
    \label{fig:SPHF-Block}
\end{figure} 

%% file: pics/sec4_pic_correction.tex
\begin{figure}[t]
    \centering
    \includegraphics[width=\linewidth]{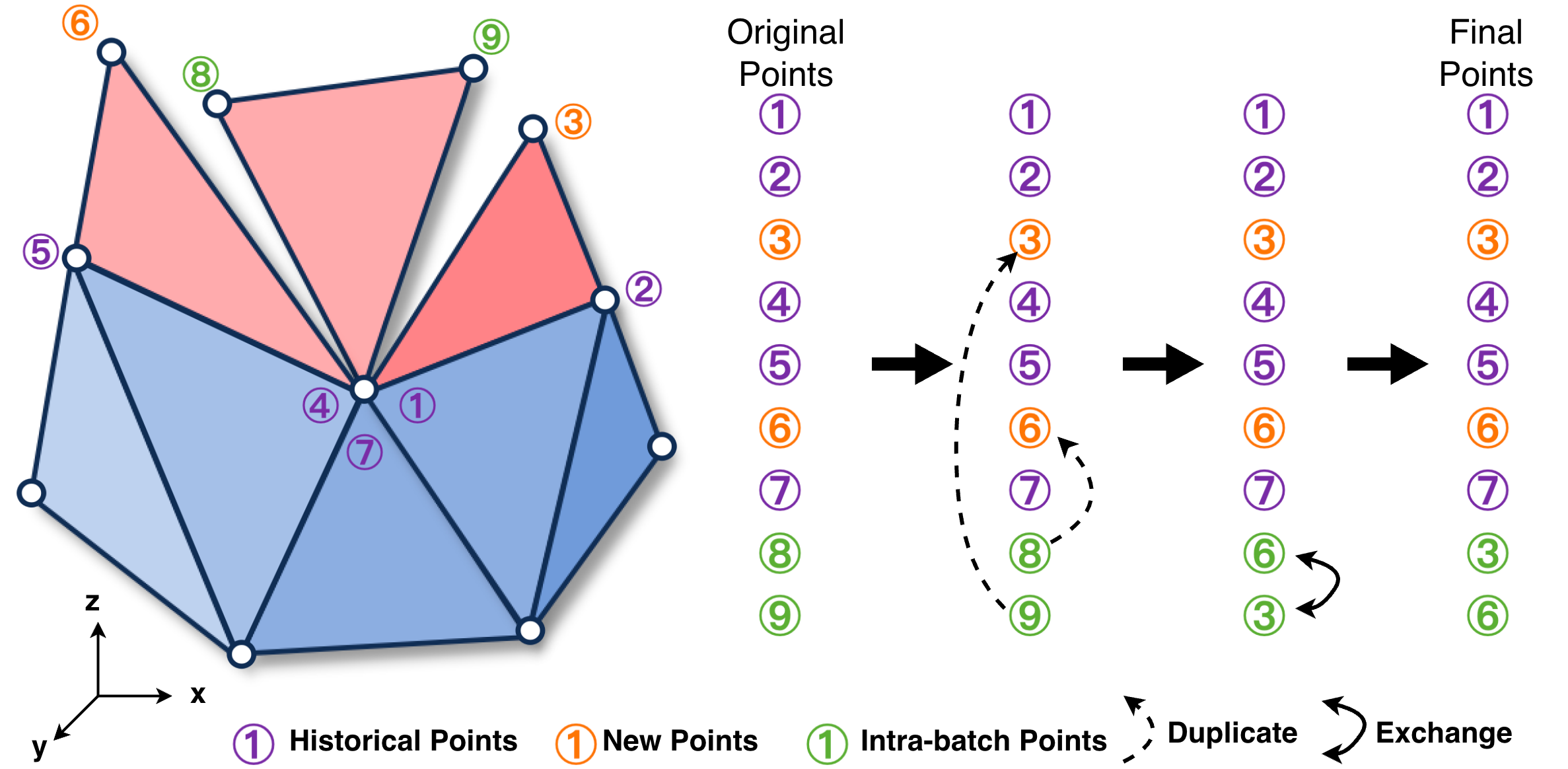}
    \caption{Example of vertex misalignment in parallel face generation and our correction mechanism.}
    \label{fig:correction}
\end{figure}

%% file: pics/sec4_pic_verify.tex
\begin{figure}[t]
    \centering
    \includegraphics[width=\linewidth]{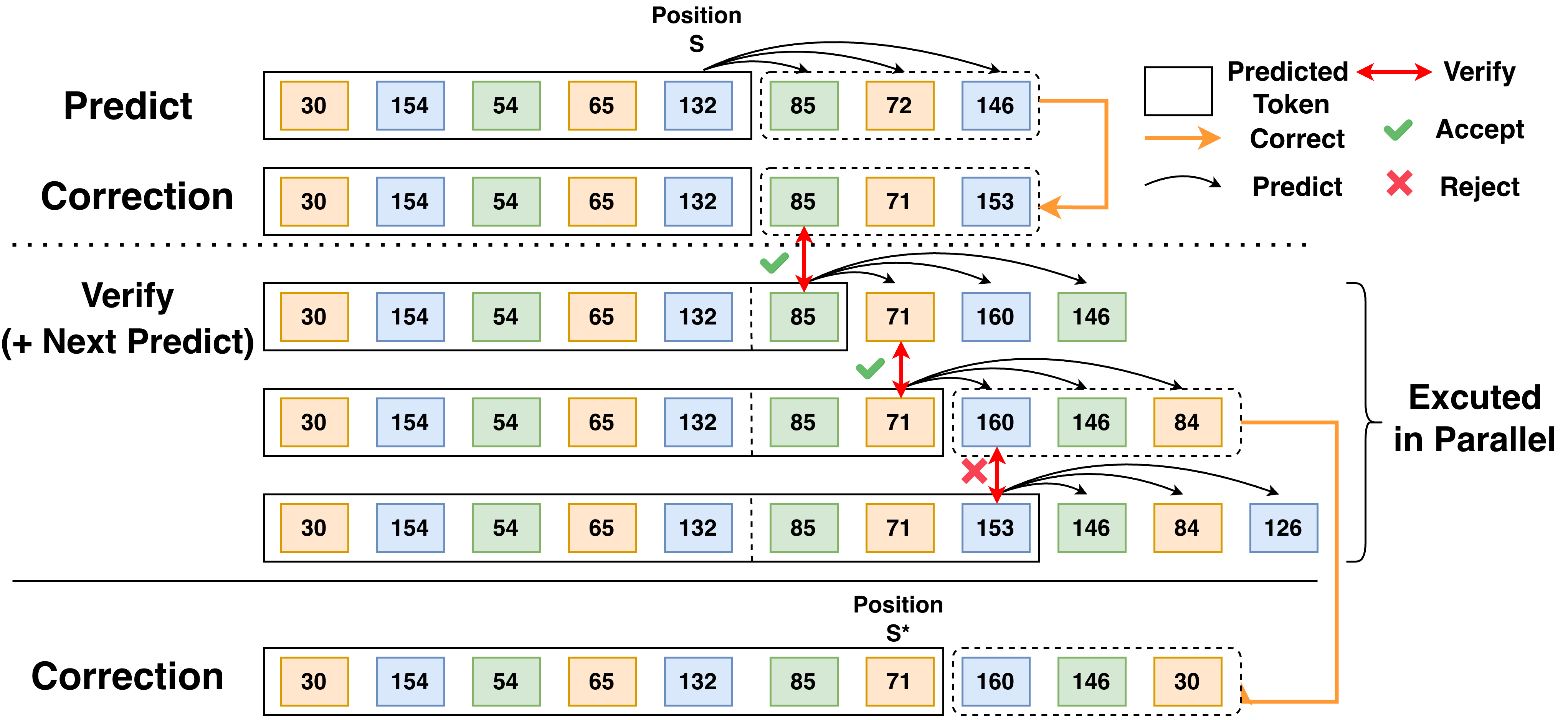}
    \caption{Example of the verify mechanism with $D=2$. The backbone verifies draft tokens in parallel, accepts consistent tokens up to $S^{*}$, and reprocesses the subsequent segment.}
    \label{fig:verify}
\end{figure}

%% file: sec/5_exp.tex
\section{Experiments}
\subsection{Experiment Settings}

\textbf{Datasets.}
The model is trained on a mixture of ShapeNetV2~\citep{chang2015shapenet}, Toys4K~\citep{stojanov2021using}, and internal data licensed from 3D content providers, comprising approximately 100K meshes without manual selection. Meshes with face lengths greater than 10,000 are filtered out. For evaluation, we select 500 meshes from ShapeNetV2 (excluded from the training set) and 500 meshes from gObjaverse~\citep{zuo2024high} as the test dataset.

\input{./pics/sec5_pic_result}
\noindent\textbf{Metrics.}
We evaluate our method using four metrics: Bounding Box IoU (bbox IoU), Chamfer Distance (CD), Hausdorff Distance (HD), Tokens per Second (TPS) and Speed-up. BBox IoU measures the overlap between the bounding boxes of generated and ground-truth meshes, reflecting global spatial consistency. CD and HD are computed from 5,000 uniformly sampled surface points of both meshes, quantifying overall geometric similarity and local reconstruction accuracy, respectively. TPS reflects the generation efficiency, defined as the total number of generated tokens divided by the total generation seconds, measured on an NVIDIA H20 GPU. Finally, we computed the actual speedup ratio (Speed-up) based on TPS.

\noindent\textbf{Baselines.}
We compare our method against four state-of-the-art autoregressive mesh generation models: BPT~\cite{weng2025scaling}, DeepMesh~\cite{zhao2025deepmesh}, Mesh-RFT~\cite{liu2025mesh}, and Meshtron~\cite{hao2024meshtron}. Since DeepMesh has only released a 0.5B-parameter configuration, we use this version for evaluation. In addition, because Mesh-RFT is not publicly available, we reproduce only its Hourglass Transformer architecture without including the M-DPO component.

\noindent\textbf{Implementary details.}
All experiments are conducted on 16 NVIDIA H20 GPUs. 
FlashMesh is implemented based on the Hourglass Transformer architecture with layer configuration 4–8–12. 
We train four variants with different capacities: 
FlashMesh (Meshtron-0.5B) uses a hidden size of 768 and a window size of 18K, trained for 3 days; 
FlashMesh (Meshtron-1B) increases the hidden size to 1536 and is trained for 5 days; 
FlashMesh (Mesh-RFT) further scales window size to 36k, trained for 5 days;
FlashMesh (Meshtron-2B) further scales the hidden size to 2048, trained for 10 days. 
The learning rate is set to $8\times10^{-5}$. 
During speculative decoding, each face-level step predicts 18 tokens and each point-level step predicts 15 tokens. 
The loss balancing factor $\gamma$ is set to 0.3. 
Additional implementation and optimization details are provided in the supplementary material.

\input{tables/sec3_main_exp.tex}

\subsection{Quantitative Results}
\cref{tab:results} reports the quantitative comparison. We first compare with two methods, BPT and DeepMesh. We do not integrate our framework to BPT or DeepMesh since both employ token compression techniques that are not compatible with our framework, and such compression also contributes to their lower generation quality.
We also experiment with two hourglass-transformer–based methods, Meshtron and Mesh-RFT. 
We integrate our framework into both baselines: ``Ours (Meshtron)'' denotes FlashMesh built on Meshtron, and ``Ours (Mesh-RFT)'' denotes FlashMesh built on Mesh-RFT. 
For completeness, we report Mesh-RFT results as well as Meshtron results for both the 1B and 2B configurations. 
The results show that FlashMesh achieves substantial runtime improvements while enhancing mesh generation quality.

\subsection{Qualitative Results}
We conduct a qualitative comparison of our method against established baselines and present several challenging examples in~\cref{fig:result}. 
Specifically, we highlight failure modes of BPT, DeepMesh, Mesh-RFT and Meshtron (2B) in terms of generation quality.
Through these qualitative examples, we demonstrate the effectiveness of our method in achieving faster and higher-quality mesh generation.

\subsection{Ablation}
\textbf{Speculative decoding and correction mechanism.}
We first compare the generation quality and acceleration performance under different configurations of the speculative decoding and correction mechanisms.
(A) The baseline is the Meshtron model with 1B parameters.
(B) On top of the baseline, we add the SP-Block. Since both the face-level and point-level decoding require the participation of the HF-Block, the SP-Block is applied only at the coordinate level.
(C) We then add both the SP-Block and the HF-Block to the baseline.
(D) Finally, we introduce the SP-Block, HF-Block, and the correction mechanism together.

\cref{tab:ablation} presents the evaluation results of these configurations.
After adding the SP-Block, the model begins speculative decoding and achieves a certain acceleration, though the improvement remains limited.
With the HF-Block incorporated, the model gains access to higher-level contextual information, enabling more accurate multi-token predictions and achieving a more significant speedup.
When the correction mechanism is further introduced, the parallel predictions are refined with higher accuracy, leading to the best acceleration while enhancing mesh quality.

\input{tables/sec3_ablation.tex}
\input{tables/sec3_draft_num.tex}

\noindent\textbf{Number of draft token in face-level and point-level.}
Increasing the number of predicted draft tokens allows the model to generate more tokens in parallel. However, as the prediction extends further into future positions, accuracy tends to decrease, and excessive token prediction also increases computational cost. Therefore, a trade-off must be made in determining the optimal number of predicted tokens to maximize acceleration. In our framework, speculative decoding occurs at least at the point or face level every three coordinate tokens. Consequently, the number of draft tokens at the point level is fixed at two. When the number of predicted tokens exceeds this value, overlapping predictions inevitably occur between coordinate-, point-, and face-level decoding, which in turn limits the acceleration efficiency. The detailed analysis of this phenomenon is provided in the supplementary material. Hence, the number of draft tokens predicted at the face and point levels directly determines the overall runtime performance. Specifically, the number of face-level draft tokens must be a multiple of nine (as one face token corresponds to nine coordinate tokens), and the number of point-level draft tokens must be a multiple of three (as one point token corresponds to three coordinate tokens). \cref{tab:draft_num} summarizes the impact of different draft token configurations on runtime efficiency. $n-m$ means the number of draft token from face-level and point-level are $n$ and $m$, respectively. Face-Acc and Point-Acc denote the average number of correctly verified tokens per draft prediction. We observe that using too few draft tokens results in limited acceleration, while predicting too many leads to accuracy degradation and much computational cost, even affecting the overall mesh generation quality.

\noindent\textbf{Parameters of model.}
We further investigate the impact of model size on generation quality and runtime efficiency. Specifically, we apply our FlashMesh framework to Meshtron models with 0.5B, 1B, and 2B parameters. \cref{tab:param} shows that larger models gain better generation quality while achieving higher acceleration efficiency. In addition, we observe a slight degradation in generation quality when the model size is very small (0.5B). Following the conclusion of~\cite{gloeckle2024better}, we attribute this to the limited representational and reasoning capacity of smaller models, which may be insufficient to support multi-token prediction without sacrificing quality.
\input{tables/sec3_params.tex}

\noindent\textbf{Loss weight $\gamma$.}
\input{tables/sec3_gamma.tex}
We further investigate the effect of the weighting coefficient $\gamma$ in the loss function. 
As shown in~\cref{tab:gamma}, the results indicate that varying this hyperparameter has minimal impact on model performance. 
Therefore, we set $\gamma = 0.3$ in our final configuration.

%% file: pics/sec5_pic_result.tex
\begin{figure*}[t]
    \centering
    \includegraphics[width=0.905\linewidth]{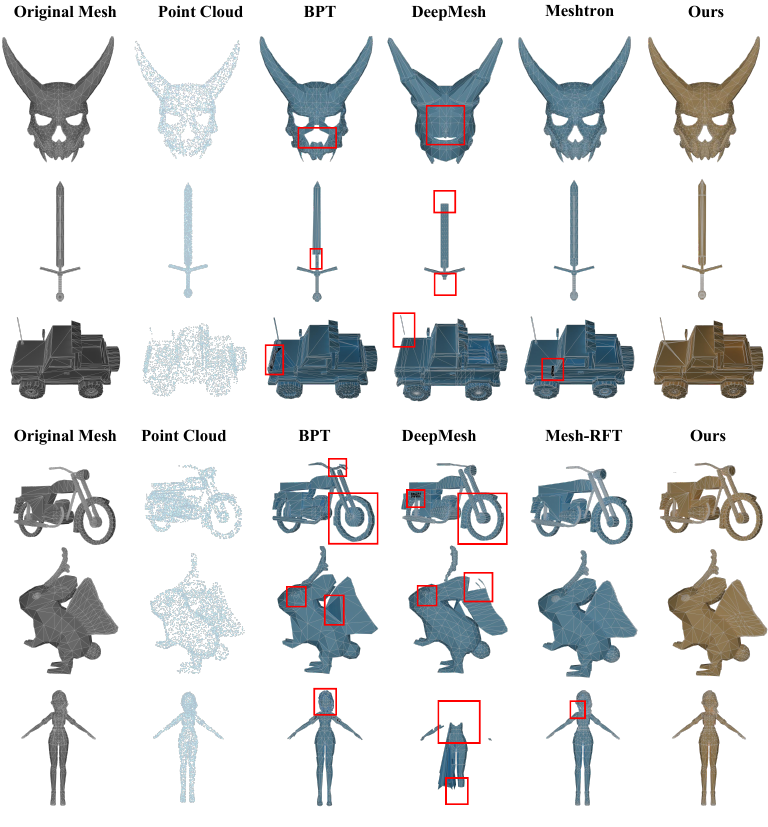}
    \vspace{-4mm}\caption{Qualitative comparison of mesh generation results. We compare FlashMesh against baseline methods including BPT and DeepMesh. Besides, in the top three samples, we also show the results of Ours (Meshtron-2B) and Meshtron-2B, while in the bottom three samples, we also present Ours (Mesh-RFT) and Mesh-RFT. Our method, FlashMesh, achieving high geometric fidelity while significantly accelerating the generation process.
    }\vspace{-4mm}
    \label{fig:result}
\end{figure*} 

%% file: tables/sec3_main_exp.tex
\begin{table}[t]
  \caption{Quantitative comparison of mesh generation methods.
  FlashMesh (Ours) achieves the best trade-off between quality (CD, HD, BBox-IoU), efficiency (TPS) and Speed-up.
  All results are measured on the H20 GPU.}\vspace{-2mm}
  \label{tab:results}
  \centering
  \resizebox{\linewidth}{!}{
  \begin{tabular}{@{}lcccccc@{}}
    \toprule
    Method & Param. (B) & CD $\downarrow$ & HD $\downarrow$ & BBox-IoU $\uparrow$ & TPS $\uparrow$ & Speed-up $\uparrow$ \\
    \midrule
    BPT~\cite{weng2025scaling}                & 0.7 & 0.128 & 0.280 & 0.894 & 29.1 & - \\
    DeepMesh~\cite{zhao2025deepmesh}          & 0.5 & 0.139 & 0.297 & 0.870 & 40.6 & - \\
    Mesh-RFT~\cite{liu2025mesh}               & 1.1 & \cellcolor{thirdcolor}{0.114} & 0.254 & 0.912 & 95.5 & - \\
    Meshtron (1B)~\cite{hao2024meshtron}      & 1.1 & 0.121 & 0.269 & 0.901 & 98.6 & - \\
    Meshtron (2B)~\cite{hao2024meshtron}      & 2.3 & \cellcolor{secondcolor}{0.092} & \cellcolor{secondcolor}{0.206} & \cellcolor{secondcolor}{0.942} & 67.3 & - \\
    \midrule
    Ours (Mesh-RFT)    & 1.6 & \cellcolor{thirdcolor}{0.114} & \cellcolor{thirdcolor}{0.252} & \cellcolor{thirdcolor}{0.913} & \cellcolor{secondcolor}{179.2} & \cellcolor{secondcolor}{$\times$ 1.87} \\
    Ours (Meshtron 1B) & 1.6 & 0.120 & 0.267 & 0.905 & \cellcolor{bestcolor}{180.4} & \cellcolor{thirdcolor}{$\times$ 1.83}\\
    Ours (Meshtron 2B) & 3.4 & \cellcolor{bestcolor}{0.089} & \cellcolor{bestcolor}{0.198} & \cellcolor{bestcolor}{0.949} & \cellcolor{thirdcolor}{136.6} & \cellcolor{bestcolor}{$\times$ 2.03}\\
    \bottomrule
  \end{tabular}
  }
\end{table}


%% file: tables/sec3_ablation.tex
\begin{table}[t]
  \caption{Ablation study on different speculative decoding and correction configurations.}\vspace{-2mm}
  \label{tab:ablation}
  \centering
  \resizebox{\linewidth}{!}{
  \begin{tabular}{@{}lcccc@{}}
    \toprule
    Configuration & CD ↓ & HD ↓ & BBox-IoU ↑ & TPS ↑ \\
    \midrule
    \textbf{A}  Meshtron 1B                      & 0.121 & 0.269 & 0.901 & 95.5 \\
    \textbf{B}  w.\,SP-Block                     & 0.122 & 0.269 & 0.903 & 109.7 \\
    \textbf{C} w.\,SP-Block\,+\,HF-Block        & 0.120 & 0.268 & 0.904 & 176.5 \\
    \textbf{D} w.\,SP-Block\,+\,HF-Block\,+\,Correction & 0.120 & 0.267 & 0.905 & 180.4 \\
    \bottomrule
  \end{tabular}
  }
\end{table} 

%% file: tables/sec3_draft_num.tex
\begin{table}[t]
  \caption{Effect of different face-level and point-level draft token numbers on generation quality and speed.
  $n\!-\!m$ in configuration means the number of draft tokens from face-level and point-level are $n$ and $m$, respectively.
  Face-Acc and Point-Acc denote the average number of correctly verified tokens per draft prediction.}\vspace{-2mm}
  \label{tab:draft_num}
  \centering
  \resizebox{\linewidth}{!}{
  \begin{tabular}{@{}lcccccc@{}}
    \toprule
    Configuration & Face-Acc & Point-Acc & CD ↓ & HD ↓ & TPS ↑ & Speed-up ↑ \\
    \midrule
    Ori Meshtron 1B & - & - & 0.121 & \cellcolor{thirdcolor}{0.269} & 98.6 & $\times$1.00 \\
    9–9     & 6.43/9  & 6.97/9  & 0.121 & 0.270 & 139.9 & $\times$1.52 \\
    27–27   & 8.24/27 & 8.39/27 & 0.127 & 0.278 & 114.4 & $\times$1.16 \\
    27–9    & 10.07/27 & 7.04/9  & 0.121 & \cellcolor{thirdcolor}{0.269} & 143.0 & $\times$1.45 \\
    18–18   & 9.84/18 & 10.35/18 & \cellcolor{bestcolor}{0.120} & \cellcolor{thirdcolor}{0.269} & \cellcolor{secondcolor}{179.9} & \cellcolor{secondcolor}{$\times$1.82} \\
    18–12   & 9.85/18 & 9.16/12 & \cellcolor{bestcolor}{0.120} & \cellcolor{bestcolor}{0.266} & \cellcolor{thirdcolor}{178.1} & \cellcolor{thirdcolor}{$\times$1.81} \\
    18–15   & 9.80/18 & 10.04/15 & \cellcolor{bestcolor}{0.120} & \cellcolor{secondcolor}{0.267} & \cellcolor{bestcolor}{180.4} & \cellcolor{bestcolor}{$\times$1.83} \\
    \bottomrule
  \end{tabular}
  }
\end{table}


%% file: tables/sec3_params.tex
\begin{table}[t]
  \caption{Quantitative comparison of different parameters.
  We conduct experiments based on 0.5B, 1B and 2B of the original Meshtron method as well as that of our FlashMesh method.
  All results are measured on the H20 GPU.}\vspace{-2mm}
  \label{tab:param}
  \centering
  \resizebox{\linewidth}{!}{
  \begin{tabular}{@{}lcccccc@{}}
    \toprule
    Method & Param. (B) & CD $\downarrow$ & HD $\downarrow$ & BBox-IoU $\uparrow$ & TPS $\uparrow$ & Speed-up $\uparrow$ \\
    \midrule
    Meshtron (0.5B)           & 0.5 & 0.137 & 0.296 & 0.852 & 112.1 & - \\
    Meshtron (1B)             & 1.1 & 0.121 & 0.269 & 0.901 & 98.6 & - \\
    Meshtron (2B)             & 2.3 & \cellcolor{secondcolor}{0.092} & \cellcolor{secondcolor}{0.206} & \cellcolor{secondcolor}{0.942} & 67.3 & - \\
    \midrule
    Ours (Meshtron 0.5B)     & 0.8 & 0.140 & 0.305 & 0.843 & \cellcolor{secondcolor}{164.4} & \cellcolor{thirdcolor}{$\times$ 1.47} \\
    Ours (Meshtron 1B)       & 1.6 & \cellcolor{thirdcolor}{0.120} & \cellcolor{thirdcolor}{0.267} & \cellcolor{thirdcolor}{0.905} & \cellcolor{bestcolor}{180.4} & \cellcolor{secondcolor}{$\times$ 1.83} \\
    Ours (Meshtron 2B)       & 3.4 & \cellcolor{bestcolor}{0.089} & \cellcolor{bestcolor}{0.198} & \cellcolor{bestcolor}{0.949} & \cellcolor{thirdcolor}{136.6} & \cellcolor{bestcolor}{$\times$ 2.03} \\
    \bottomrule
  \end{tabular}
  }\vspace{-3mm}
\end{table}


%% file: tables/sec3_gamma.tex
\begin{table}[t]
  \caption{Ablation study on different values of $\gamma$.}\vspace{-2mm}
  \label{tab:gamma}
  \centering
  \resizebox{0.7\linewidth}{!}{
  \begin{tabular}{@{}lcccc@{}}
    \toprule
    The value of $\gamma$ & CD ↓ & HD ↓ & BBox-IoU ↑ & TPS ↑ \\
    \midrule
    0.1                      & 0.120 & 0.267 & 0.905 & 180.3 \\
    0.3                    & 0.120 & 0.267 & 0.905 & 180.4 \\
    0.5       & 0.120 & 0.269 & 0.904 & 180.1 \\
    \bottomrule
  \end{tabular}
  }\vspace{-6mm}
\end{table} 

%% file: sec/6_conclusion.tex
\section{Conclusion}
\label{sec:conclusion}

In this paper, we presented FlashMesh, an efficient and high-quality framework for autoregressive mesh generation. By introducing a predict–correct–verify paradigm, FlashMesh enables parallel multi-token prediction while maintaining geometric consistency and topological fidelity. Our speculative decoding strategy, combined with a structure-aware correction mechanism, significantly improves both inference speed and generation quality. Experiments validate that FlashMesh outperforms existing baselines in both efficiency and fidelity. FlashMesh offers a promising step toward fast, scalable 3D content creation in practical applications.

\noindent\textbf{Limitations and Future Work.} While FlashMesh substantially accelerates mesh generation, it still inherits the inherent limitations of autoregressive models, such as sensitivity to early prediction errors. In future work, we aim to explore hybrid decoding strategies and integrate geometric priors more explicitly to further enhance robustness.

%% file: sec/X_suppl.tex
\clearpage

\maketitlesupplementary
\newcommand{\FlashMesh}{\textsc{FlashMesh}}


\begin{strip}
    \centering
    \includegraphics[width=\textwidth]{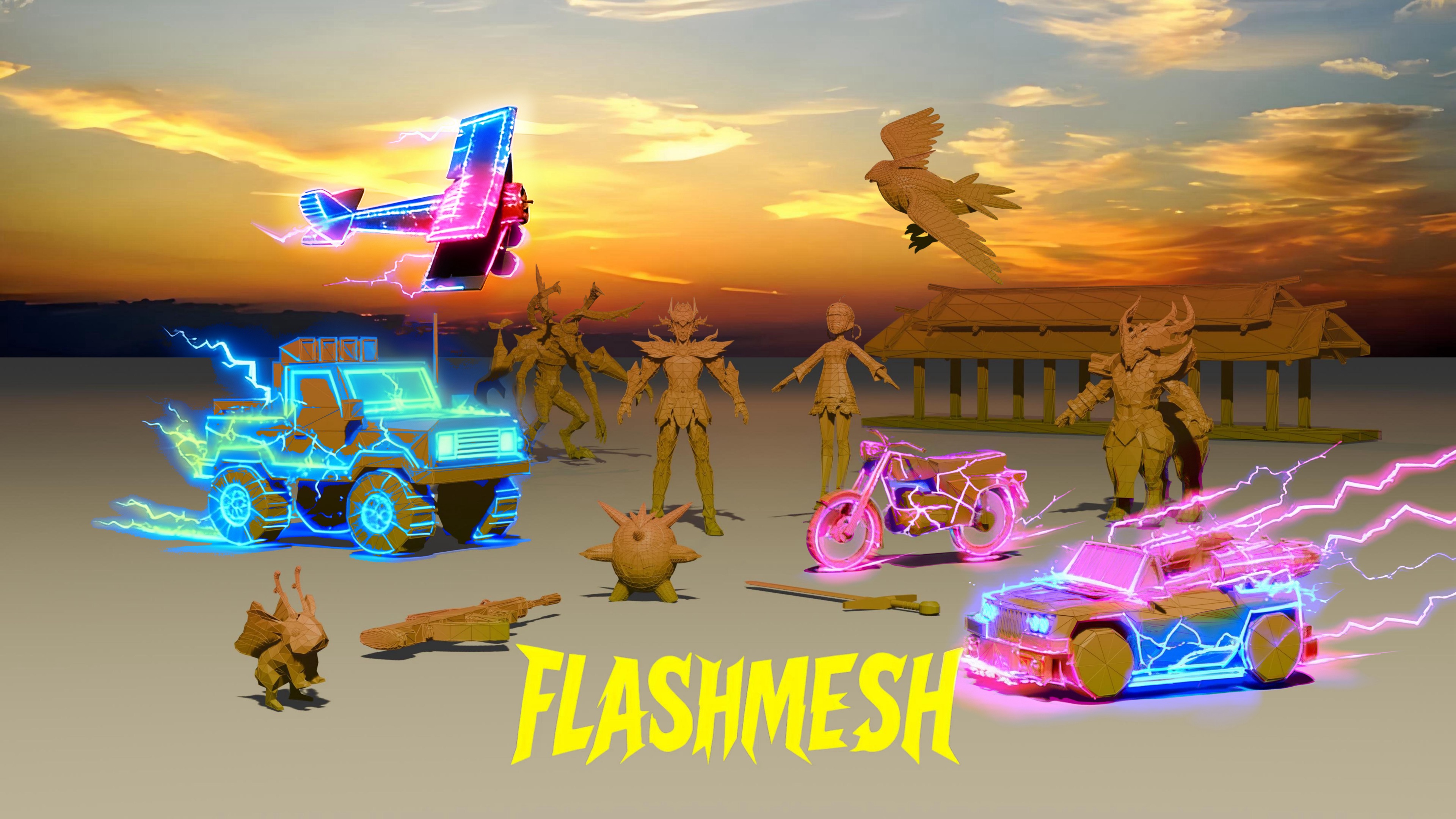}
    \captionof{figure}{Artistic meshes generated by \FlashMesh.}
    \label{fig:supp_teaser}
\end{strip}

\section{Supplementary Material Overview}

This supplementary document provides additional technical and empirical details to support the main paper. An overview of artistic meshes generated by FlashMesh is shown in Figure~\ref{fig:supp_teaser}, demonstrating its capability to produce diverse, high-fidelity 3D content. The supplementary is organized into five sections:

\paragraph{\cref{sec: model} Model Architecture.} We provide detailed explanations of the hourglass Transformer, node splitting strategy, and the hierarchical three-level speculative decoding process. We also describe the optimization techniques used in our methods.

\paragraph{\cref{sec: work} Why Does It Work? Theoretical Insights.} We present concise mathematical analysis to explain how structured multi-token speculation improves both speed and generation quality.

\paragraph{\cref{sec: data} Dataset Statistics.} We summarize the mesh data distribution used for training, including the face-count statistics after filtering out overly complex samples.

\paragraph{\cref{sec: train} Training Configuration.} We provide full details on training schedules, hyperparameters, and optimization strategies for all model variants.

\paragraph{\cref{sec: result} Additional Results.} We include extended qualitative results showing that FlashMesh consistently generates more coherent and visually accurate meshes compared to baseline methods.

\input{./sec/supp1_model}
\input{./sec/supp2_proof}
\input{./sec/supp3_data}
\input{./sec/supp4_train_details}
\input{./sec/supp5_result}

%% file: sec/supp1_model.tex
\section{Model Details}
\label{sec: model}
\input{./pics/model}
\subsection{Hourglass Transformer}

Hourglass Transformer is designed to explicitly model the hierarchical structure of meshes across three levels: faces, vertices, and coordinates. As shown in the left panel of ~\cref{fig:supp_model}, the network begins with a sequence of coordinate tokens, where every three coordinates form a vertex, and every three vertices (i.e., nine tokens) define a triangular face. The hourglass architecture processes these tokens through successive shortening layers, which compress groups of tokens into higher-level representations: coordinate embeddings are aggregated into vertex embeddings, and vertex embeddings are further compressed into face embeddings. These compact representations capture coarse geometric and topological information. Then, through corresponding upsampling layers, the model expands face-level embeddings back to vertex- and coordinate-level sequences, progressively refining details while maintaining global consistency. Residual connections between levels enable information flow across different resolutions. This hierarchical compression–expansion mechanism allows the model to efficiently capture both local geometric relationships and global structural dependencies, leading to improved quality and efficiency compared to conventional flat transformer decoders.

\subsection{Split Nodes}
Within the hourglass hierarchy, the upsampling split nodes serve as transition points where higher-level features are expanded into finer-grained mesh representations. They are located between the up/down-sampling layers and the Transformer layers. Specifically, the split nodes fall into three categories: face split nodes, point split nodes, and coordinate split nodes, which are depicted in~\cref{fig:supp_model} (left) as red, pink, and orange squares, respectively.

\subsection{Three-Level Speculative Decoding}

We now introduce how the three types of split nodes generate multiple future tokens through the SP-Block and HF-Block, and how their predictions are merged to produce draft tokens.

\textbf{Face-level speculative decoding.}
When the token passes through the face-level split nodes, it is processed by one SP-Block and two HF-Blocks, enabling multi-token prediction. As illustrated by the red block in~\cref{fig:supp_model}~(middle), the face split nodes first go through a transformer layer and the SP-Block to produce several future face tokens. These predictions are then refined by two HF-Blocks, which interact with the previously generated backbone tokens, ultimately producing multiple future draft tokens.

\textbf{Point-level speculative decoding.}
When the token reaches the point-level split nodes, it is processed by one SP-Block followed by one HF-Block. As shown by the pink block in~\cref{fig:supp_model}~(right), the point split nodes generate several future point tokens after passing through the transformer and SP-Block. These predictions are then refined by a single HF-Block and interact with the backbone tokens to produce additional future draft tokens.

\textbf{Coordinate-level speculative decoding.}
When the token reaches the coordinate-level split nodes, it passes only through one SP-Block. As indicated by the orange block in~\cref{fig:supp_model}~(right), the coordinate split nodes directly generate one main token through transformer and several future draft tokens through SP-Block.

\textbf{Integrating the three levels.}
During a single forward pass, multiple types of split nodes may be triggered, resulting in multiple predictions for the same future position. To address this, we average the predicted distributions from all active levels and treat the averaged distribution as the final prediction for that position. In addition, some forward passes may not encounter face- or vertex-level split nodes, but we can still reuse the most recently generated draft tokens from those levels.

\subsection{Optimizations}

We observe that a single position may be predicted by multiple levels of speculative decoding, which leads to redundant computation and reduced efficiency. Our goal is to allow each position to be predicted by exactly one level whenever possible. We introduce two optimization rules:

(A) \textit{Remove point-level predictions immediately following face-level predictions.}
As shown on the left side of~\cref{fig:supp_model}, the row corresponding to $S_8$ receives both face-level and point-level predictions. Because their prediction ranges heavily overlap, this causes unnecessary duplicated computation. We therefore remove the point-level predictions in such cases and keep the higher-level face predictions.

(B) \textit{Remove the first three draft token predictions from face- and point-level decoding.}
We observe that the predictions made at the coordinate level overlap substantially with those from the point- and face-levels. To reduce duplication, we remove the first point token (i.e. the first three coordinate tokens) predictions from face-level and point-level decoding, and let the coordinate-level modules exclusively predict the first three future tokens for each position. Alternatively, we could remove the first two point tokens and let the coordinate-level modules exclusively predict the first six future tokens for each position. However, since every input token must pass through the coordinate-level modules, this approach would significantly increase the number of predicted tokens and thereby reduce the overall inference efficiency.

Together, these optimizations significantly reduce redundant predictions and improve model efficiency.

We illustrate two representative steps using a 9-9 FlashMesh model:
(1) When predicting $S_5$, the point-level modules output $\tilde{S}_8$–$\tilde{S}_{13}$, while the coordinate-level SP-Block outputs $\tilde{S}_6$ and $\tilde{S}_7$, and the backbone outputs $\tilde{S}_5$. In total, we obtain one main token and eight draft tokens.
(2) When predicting $S_6$, the coordinate-level SP-Block outputs $\tilde{S}_7$ and $\tilde{S}_8$, and the backbone outputs $\tilde{S}_6$. We also reuse the previously generated point-level predictions $\tilde{S}_8$–$\tilde{S}_{13}$, averaging the two predictions for $\tilde{S}_8$. This produces one main token and seven draft tokens.

\subsection{Experiments}

\textbf{Ablation on optimization rules.}
\input{./tables/optimization}
We conduct ablations to evaluate the effect of the two optimization rules. As shown in~\cref{tab:supp_optimization}, none of the optimizations degrade generation quality, and both of them contribute to improved inference speed.

\textbf{Ablation on multi-level speculative decoding.}
\input{./tables/abliation}
A simple baseline is to remove the face- and point-level modules and directly predict a large number of tokens at the coordinate level. To validate the advantage of our hierarchical three-level design, we implement this variant of predicting 18 tokens at the coordinate level.~\cref{tab:supp_ablation} shows that our multi-level approach achieves superior speedup, as the variant generates roughly twice as many tokens, which slows down inference.

%% file: pics/model.tex
\begin{figure*}[t]
    \centering
    \includegraphics[width=\linewidth]{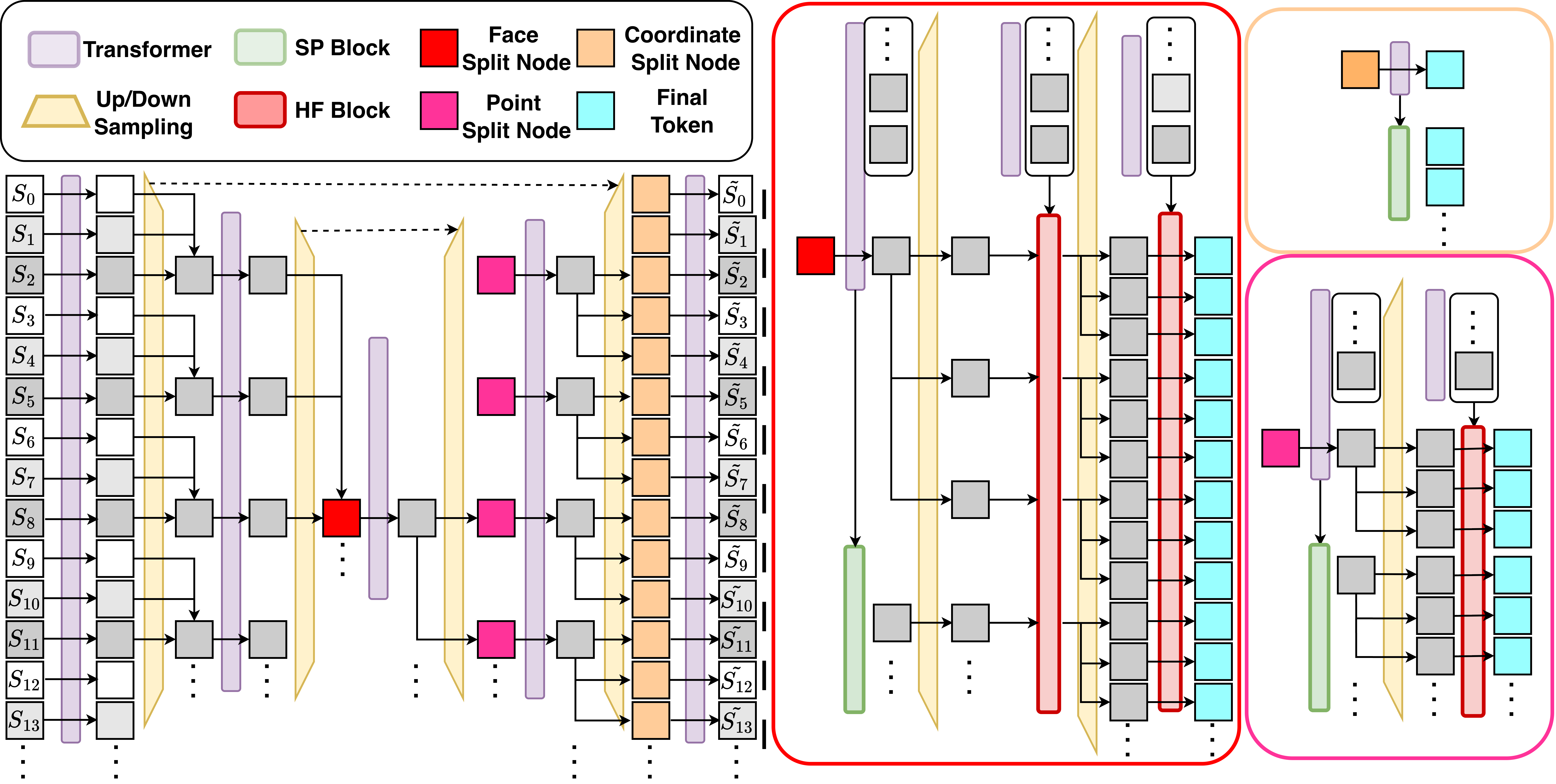}
    \caption{Overall architecture of the multi-layer multi-head speculative decoding. Left: original hourglass transformer. Red box in the middle: face-level pipeline of multi-layer multi-head speculative decoding. Pink box on the right: point-level pipeline of multi-layer multi-head speculative decoding. Orange box on the right: coordinate-level pipeline of multi-layer multi-head speculative decoding.}
    \label{fig:supp_model}
\end{figure*} 

%% file: tables/optimization.tex
\begin{table}[t]
  \caption{Ablation study on different optimizations.}\vspace{-2mm}
  \label{tab:supp_optimization}
  \centering
  \resizebox{\linewidth}{!}{
  \begin{tabular}{@{}lcccc@{}}
    \toprule
    Configuration & CD ↓ & HD ↓ & BBox-IoU ↑ & TPS ↑ \\
    \midrule
    \textbf{A}  w/o any optimizations                      & 0.120 & 0.267 & 0.905 & 176.0 \\
    \textbf{B} w.\, (A)        & 0.120 & 0.267 & 0.905 & 178.3 \\
    \textbf{C} w.\, (A) + (B) & 0.120 & 0.267 & 0.905 & 180.4 \\
    \bottomrule
  \end{tabular}
  }
\end{table} 

%% file: tables/abliation.tex
\begin{table}[t]
  \caption{Ablation study on the variant.}\vspace{-2mm}
  \label{tab:supp_ablation}
  \centering
  \resizebox{0.8\linewidth}{!}{
  \begin{tabular}{@{}lcccc@{}}
    \toprule
    Configuration & CD ↓ & HD ↓ & BBox-IoU ↑ & TPS ↑ \\
    \midrule
    \textbf{A}  Variant                     & 0.120 & 0.268 & 0.906 & 166.1 \\
    \textbf{B}  Ours               & 0.120 & 0.267 & 0.905 & 180.4 \\
    \bottomrule
  \end{tabular}
  }
\end{table} 

%% file: sec/supp2_proof.tex
\section{Why Does It Work? Theoretical Insights.}
\label{sec: work}

Why can our \textit{structured speculating} method both accelerate inference and improve generation quality? We provide explanations from two complementary perspectives: speed and quality. Specifically, ~\cref{sec: fast} derives the expected speedup formula and explains the underlying computational advantage, while ~\cref{sec: quality}  analyzes the quality improvement from an information-theoretic viewpoint.

\subsection{Why It Is Fast: A Computational Efficiency Analysis}
\label{sec: fast}

A standard autoregressive model consumes one token and predicts the next-token distribution, generating outputs strictly one-by-one. This inherently sequential process leads to slow inference. As described in the main paper, our method employs lightweight SP-Blocks and HF-Blocks to predict multiple future tokens in parallel. After the prediction, a correction and verification stage determines how many of these draft tokens can be safely accepted, thus reducing the total number of forward passes.

We find that the effective speedup is governed by several factors: the original per-token generation latency of the baseline model ($T_{\mathrm{ori}}$), the cost of our model when consuming $n$ input tokens and speculating $k$ future tokens ($T^{n,k}_{\mathrm{ours}}$), and the average number of accepted tokens $m$ per speculation cycle. The overall speedup ratio $\mathcal{S}$ can be expressed as:

\begin{equation}
\mathcal{S} = \frac{T_{\mathrm{ori}} \cdot m}{T^{n,k}_{\mathrm{ours}}}.
\end{equation}

These observations indicate that the speedup increases when (1) the average number of accepted tokens $m$ grows, (2) the speculative forward pass remains lightweight, and (3) the ratio between accepted tokens and computation cost improves. In practice, our structured design keeps the speculative forward pass extremely lightweight, enabling scalable acceleration across different model sizes.

\subsection{Why It Has Better Quality: An Information-Theoretic Argument}
\label{sec: quality}

We now give a compact information-theoretic argument explaining why structured multi-token speculation improves generation quality in the mesh-generation setting. Let each token denote a discrete mesh-generation unit (for example: a face token, a vertex token, or a coordinate token). Consider two successive future tokens denoted by random variables \(X\) and \(Y\) (e.g. the tokens that encode two adjacent faces or two consecutive vertex coordinates). All probabilities below are implicitly conditioned on the observed context \(C\) (previously generated or given mesh context); we omit \(C\) from the notation for brevity.

A standard next-token objective trains the model to minimize the (cross-entropy) loss for \(X\) alone, which is driven by the entropy \(H(X)\). In contrast, a 2-token (multi-token) objective minimizes the joint loss for \((X,Y)\), driven by \(H(X,Y)=H(X)+H(Y\mid X)\). To compare these quantities and make explicit the role of dependencies between \(X\) and \(Y\), we use the mutual information \(I(X;Y)\) and the conditional entropies. The following equalities hold:

\begin{align}
H(X) &= H(X\mid Y) + I(X;Y), \label{eq:decomp1}\\
H(X)+H(Y) &= H(X\mid Y) + 2I(X;Y) + H(Y\mid X). \label{eq:decomp2}
\end{align}

Equation~\eqref{eq:decomp1} simply decomposes the marginal uncertainty of \(X\) into the uncertainty of \(X\) given \(Y\) plus the mutual information. Equation~\eqref{eq:decomp2} follows by writing \(H(Y)=H(Y\mid X)+I(X;Y)\) and substituting.

Now inspect what changes when the training / decoding objective moves from single-step prediction (minimizing terms proportional to \(H(X)\)) to multi-step prediction (minimizing terms proportional to \(H(X)+H(Y)\)). Comparing \eqref{eq:decomp1} and \eqref{eq:decomp2} shows that the multi-token objective places a larger emphasis on the mutual-information term \(I(X;Y)\): in the two-token objective \(I(X;Y)\) appears with coefficient \(2\), whereas in the single-token objective it effectively appears with coefficient \(1\). Intuitively, multi-token prediction amplifies the contribution of pairwise (and, by extension, higher-order) dependencies among neighboring tokens.

Why does this matter for mesh generation? In meshes, nearby tokens are strongly coupled: adjacent faces share vertices; vertex coordinate tokens enforce geometric continuity; semantic groups of faces follow global shape constraints. These couplings imply substantial mutual information between tokens that are nearby in the generation order. By increasing the effective weight on mutual information during learning / speculation, the multi-token objective encourages the model to (a) allocate more representational capacity to capture dependencies across multiple future tokens, and (b) reduce uncertainty about tokens that are informative for the future structure of the mesh.

Formally, suppose the model capacity and training procedure allow a reduction in the joint entropy \(H(X,Y)\) relative to optimizing only \(H(X)\). Since
\[
H(X,Y) = H(X) + H(Y\mid X),
\]
a reduction in joint entropy implies either a reduction in \(H(X)\) (per-token uncertainty) or in \(H(Y\mid X)\) (better modeling of how \(Y\) depends on \(X\)), or both. Because mesh coherence is largely enforced by such conditional relations (shared vertices, consistent normals, edge-adjacency), improving \(H(Y\mid X)\) directly reduces structural errors that would otherwise accumulate when tokens are generated strictly one-by-one.

A final, practical point connects this information view to our \textit{structured speculating} pipeline: speculation that produces multiple draft tokens, then corrects and accepts a subset effectively trains and evaluates the model on multi-step predictions at inference time. This alignment between the model's training signal (which emphasizes joint / multi-token objectives) and the inference behavior reduces exposure bias and the compounding of short-term errors. The net result is a lower expected per-token error in geometrically relevant quantities (shared-vertex indices, coordinate continuity, face-consistency), which manifests as visibly improved mesh quality.

In summary, multi-token structured speculation amplifies the learning signal for mutual dependencies among adjacent mesh tokens, lowers joint and conditional entropies relevant to geometric consistency, and thus yields more coherent and accurate mesh generations compared to purely single-step autoregressive decoding.

%% file: sec/supp3_data.tex
\section{Data Distribution}
\label{sec: data}
\input{./pics/data}

For mesh generation models, the quality and statistical distribution of the training data have a substantial impact on model performance. To address this, we filtered out meshes with more than 10,000 faces and analyzed the face-count distribution of the training set. As shown in~\cref{fig:data}, approximately 55\% of the meshes contain fewer than 3,000 faces, about 27\% fall within the range of 3,000 to 6,000 faces, and roughly 18\% contain between 6,000 and 10,000 faces. This strategy removes overly complex meshes and helps maintain stable and reliable model performance.

%% file: pics/data.tex
\begin{figure}[t]
    \centering
    \includegraphics[width=\linewidth]{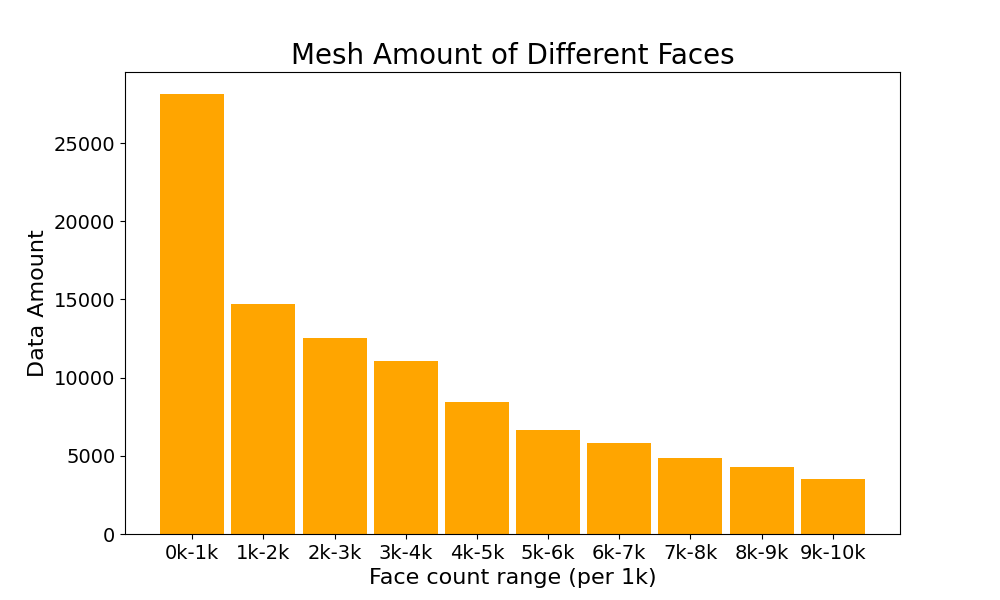}
    \caption{Data distribution: mesh amount of different faces.}
    \label{fig:data}
\end{figure} 

%% file: sec/supp4_train_details.tex
\section{More Training Details}
\label{sec: train}
\input{tables/configs2}

We evaluate our FlashMesh framework on Meshtron (0.5B)~\cite{hao2024meshtron}, Meshtron (1B)~\cite{hao2024meshtron}, Mesh-RFT~\cite{liu2025mesh}, and Meshtron (2B)~\cite{hao2024meshtron}. The detailed training configurations of these models are summarized in~\cref{tab:configs}. Specifically, we report each model's parameter count, architecture, batch size, number of layers, number of heads, dimension of the model, learning rate, and total training time. All models are trained for 30k iteration steps to ensure convergence. In addition, we adopt the Adam optimizer ($\beta_1 = 0.9$, $\beta_2 = 0.99$). In our setting, we use gradient accumulation with an accumulation step of 8, meaning that gradients from 8 mini-batches (each of size 8) are accumulated before one update, resulting in an effective batch size of 64. Moreover, we employ FlashAttention and ZeRO-2 to reduce GPU memory consumption.

%% file: tables/configs2.tex
\begin{table}[t]
  \caption{Training configurations for different models.}
  \label{tab:configs}
  \centering
  \resizebox{\linewidth}{!}{
    \begin{tabular}{@{}lcc@{}}
      \toprule
      \textbf{Setting} & \textbf{Ours (Meshtron 0.5B)} & \textbf{Ours (Meshtron 1B)} \\
      \midrule
      Parameter count & 0.8B & 1.6B \\
      Architecture    & 4-8-12 & 4-8-12 \\
      Batch size      & 64 & 64 \\
      Layers          & 24 & 24 \\
      Heads           & 8 & 16 \\
      Dimension       & 768 & 1536 \\
      Learning rate   & 1e-4 & 1e-4 \\
      Training time   & $\sim$3 days & $\sim$5 days \\
      \bottomrule

      \\[-2mm]  

      \toprule
      \textbf{Setting} & \textbf{Ours (Mesh-RFT)} & \textbf{Ours (Meshtron 2B)} \\
      \midrule
      Parameter count & 1.6B & 3.4B \\
      Architecture    & 4-8-12 & 4-10-18 \\
      Batch size      & 64 & 64 \\
      Layers          & 24 & 32 \\
      Heads           & 16 & 16 \\
      Dimension       & 1536 & 2048 \\
      Learning rate   & 1e-4 & 1e-4 \\
      Training time   & $\sim$5 days & $\sim$10 days \\
      \bottomrule
    \end{tabular}
  }
\end{table}


%% file: sec/supp5_result.tex
\section{More Results}
\label{sec: result}
We further collected more examples and presented the generation results of Meshtron (2B) and our FlashMesh (Meshtron 2B) in~\cref{fig:result1} and ~\cref{fig:result2}. Across these complex cases, our model consistently produces meshes with coherent shapes and correct topology, whereas Meshtron sometimes yields results of noticeably lower quality. This demonstrates that our method not only accelerates inference but also improves generation quality.

\input{./pics/result1}
\input{./pics/result2}

%% file: pics/result1.tex
\begin{figure*}[t]
    \centering
    \includegraphics[width=\linewidth]{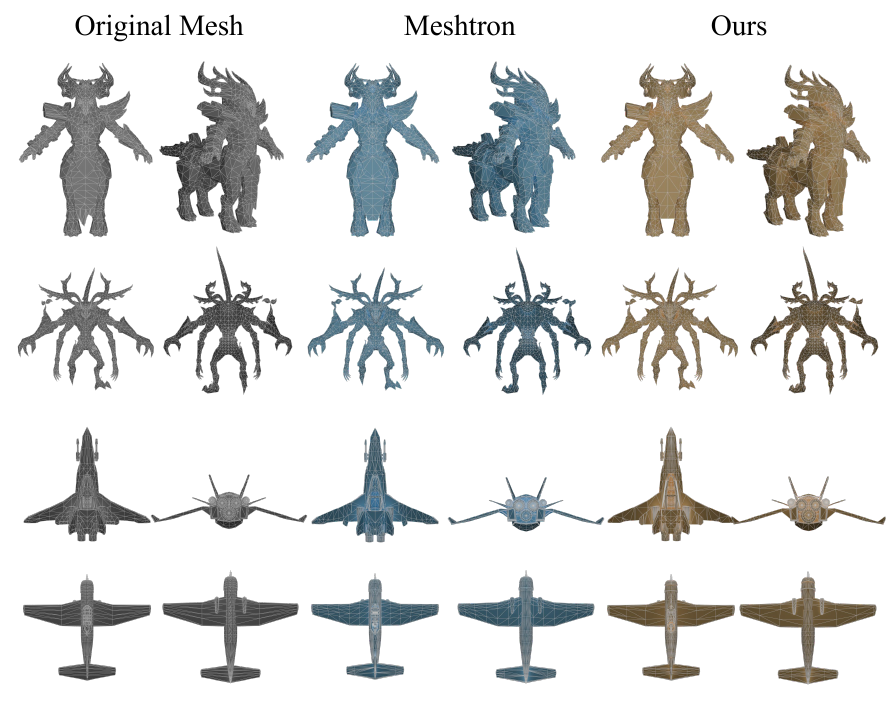}
    \caption{More results of Meshtron and Flashmesh. We present more high-fidelity results generated by our method.
    }
    \label{fig:result1}
\end{figure*} 

%% file: pics/result2.tex
\begin{figure*}[t]
    \centering
    \includegraphics[width=\linewidth]{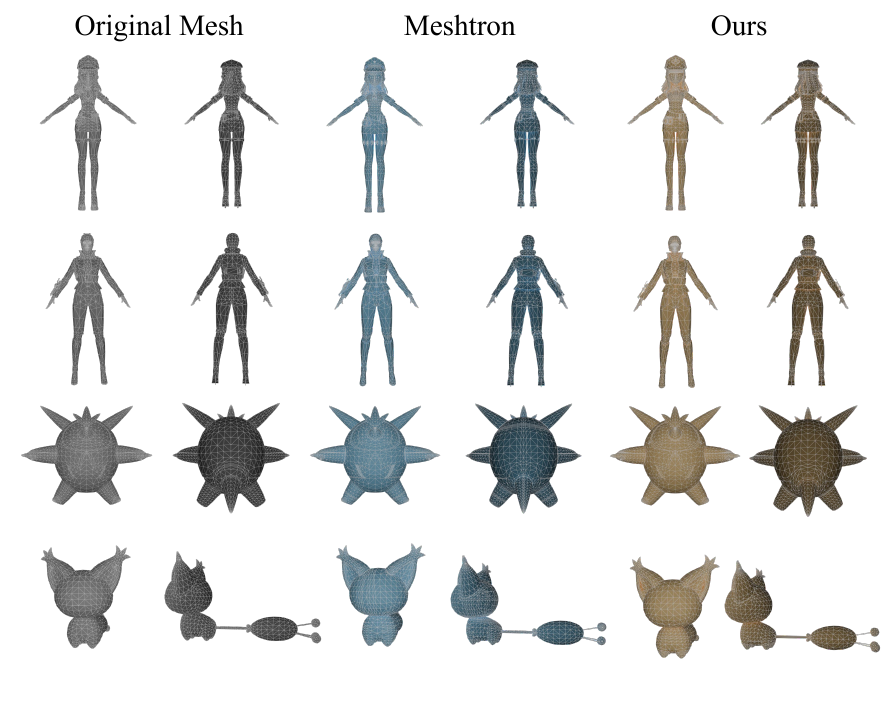}
    \caption{More results of Meshtron and Flashmesh. We present more high-fidelity results generated by our method.
    }
    \label{fig:result2}
\end{figure*}